\documentclass[11pt,a4paper]{article}
\usepackage[hyperref]{acl2020}
\usepackage{times}
\usepackage{latexsym}
\usepackage{graphicx}

\usepackage{microtype}

\aclfinalcopy 

\title{TRANS-BLSTM: Transformer with Bidirectional LSTM for Language Understanding}

\author{Zhiheng Huang \\
  Amazon AWS AI \\
  \texttt{zhiheng@amazon.com} \\\And
  Peng Xu \\
  Amazon AWS AI \\
  \texttt{pengx@amazon.com} \\\And
  Davis Liang \\
  Amazon AWS AI \\
  \texttt{liadavis@amazon.com} \\\AND
  Ajay Mishra \\
  Amazon AWS AI \\
  \texttt{misaja@amazon.com} \\\And
  Bing Xiang \\
  Amazon AWS AI \\
  \texttt{bxiang@amazon.com} \\  
  }

\date{}

\begin{document}
\maketitle
\begin{abstract}
Bidirectional Encoder Representations from Transformers (BERT) has recently achieved state-of-the-art performance on a broad range of NLP tasks including sentence classification, machine translation, and question answering. The BERT model architecture is derived primarily from the transformer. Prior to the transformer era, bidirectional Long Short-Term Memory (BLSTM) has been the dominant modeling architecture for neural machine translation and question answering. In this paper, we investigate how these two modeling techniques can be combined to create a more powerful model architecture. We propose a new architecture denoted as Transformer with BLSTM (TRANS-BLSTM) which has a BLSTM layer integrated to each transformer block, leading to a joint modeling framework for transformer and BLSTM. We show that TRANS-BLSTM models consistently lead to improvements in accuracy compared to BERT baselines in GLUE and SQuAD 1.1 experiments. Our TRANS-BLSTM model obtains an F1 score of $94.01\%$ on the SQuAD 1.1 development dataset, which is comparable to the state-of-the-art result.
\end{abstract}

\section{Introduction}
Learning representations \cite{mikolov2013} of natural language and language model pre-training \cite{devlin2018,radford2019language} has shown promising results recently. These pre-trained models serve as generic up-stream models and they can be used to improve down-stream applications such as natural language inference, paraphrasing, named entity recognition, and question answering. The innovation of BERT \cite{devlin2018} comes from the ``masked language model'' with a pre-training objective, inspired by the Cloze task \cite{taylor1953}. The masked language model randomly masks some of the tokens from the input, and the objective is to predict the original token based only on its context. 

Follow-up work including RoBERTa \cite{liu2019-2} investigated hyper-parameter design choices and suggested longer model training time. In addition, XLNet \cite{yang2019} has been proposed to address the BERT pre-training and fine-tuning discrepancy where masked tokens were found in the former but not in the latter. Nearly all existing work suggests that a large network is crucial to achieve the state-of-the-art performance. For example, \cite{devlin2018} has shown that across natural language understanding tasks, using larger hidden layer size, more hidden layers, and more attention heads always leads to better performance. However, they stop at a hidden layer size of 1024. ALBERT \cite{lan2019} showed that it is not the case that simply increasing the model size would lead to better accuracy.  In fact, they observed that simply increasing the hidden layer size of a model such as BERT-large can lead to significantly worse performance. On the other hand, model distillation \cite{hinton2015,tang2019,sun2019,sanh2019} has been proposed to reduce the BERT model size while maintaining high performance.

In this paper, we attempt to improve the performance of BERT via architecture enhancement. BERT is based on the encoder of the transformer model \cite{aswani2017}, which has been proven to obtain state-of-the-art accuracy across a broad range of NLP applications \cite{devlin2018}. Prior to BERT, bidirectional LSTM (BLSTM) has dominated sequential modeling for many tasks including machine translation \cite{chiu2016} and speech recognition \cite{graves2013}. Given both models have demonstrated superior accuracy on various benchmarks, it is natural to raise the question whether a combination of the transformer and BLSTM can outperform each individual architecture. In this paper, we attempt to answer this question by proposing a transformer BLSTM joint modeling framework. Our major contribution in this paper is two fold: 1) We propose the TRANS-BLSTM model architectures, which combine the transformer and BLSTM into one single modeling framework, leveraging the modeling capability from both the transformer and BLSTM. 2) We show that the TRANS-BLSTM models can effectively boost the accuracy of BERT baseline models on SQuAD 1.1 and GLUE NLP benchmark datasets.

\section{Related work}

\subsection{BERT}
Our work focuses on improving the transformer architecture \cite{aswani2017}, which motivated the recent breakthrough in language representation, BERT \cite{devlin2018}. Our work builds on top of the transformer architecture, integrating each transformer block with a bidirectional LSTM \cite{sepp1997}. Related to our work, XLNet \cite{yang2019} proposes two-stream self-attention as opposed to single-stream self-attention used in classic transformers. With two-stream attention, XLNet can be treated as a general language model that does not suffer from the pretrain-finetune discrepancy (the mask tokens are seen during pretraining but not during finetuning) thanks to its autoregressive formulation. Our method overcomes this limitation with a different approach, using single-stream self-attention with an integrated BLSTM layer for each transformer layer.

\subsection{Bidirectional LSTM}
The LSTM network \cite{sepp1997} has demonstrated powerful modeling capability in sequential learning tasks including named entity tagging \cite{huang2015,chiu2016}, machine translation \cite{bahdanau2015,wu2016} and speech recognition \cite{graves2013,sak2014}. The motivation of this paper is to integrate bidirectional LSTM layers to the transformer model to further improve transformer performance. The work of \cite{tang2019} attempts to distill a BERT model to a single-layer bidirectional LSTM model. It is relevant to our work as both utilizing bidirectional LSTM. However, their work leads to inferior accuracy compared to BERT baseline models. Similar to their observation, we show that in our experiments, the use of BLSTM model alone (even with multiple stacked BLSTM layers) leads to significantly worse results compared to BERT models. However, our proposed joint modeling framework, TRANS-BLSTM, is able to boost the accuracy of the transformer BERT models. 

\subsection{Combine Recurrent Network and Transformer}
Previous work has explored the combination of the recurrent network and transformer. For example, \cite{	lei2018} has substituted the feedforward network in transformer with the simple recurrent unit (SRU) implementation and achieved better accuracy in machine translation. It is similar to one of the proposed models in this paper. However, the difference is that our paper investigates the gain of the combination in BERT pre-training context, while their paper focused on the parallelization speedup of SRU in machine translation encoder and decoder context. 

\section{TRANS and Proposed TRANS-BLSTM Architectures}

In this section, we first review the transformer architecture, then propose the transformer bidirectional LSTM network architectures (TRANS-BLSTM), which integrates the BLSTM to either the transformer encoder or decoder. 

\subsection{Transformer architecture (TRANS)}
\begin{figure}[!hbt]
    \centering
    \includegraphics[scale=0.6]{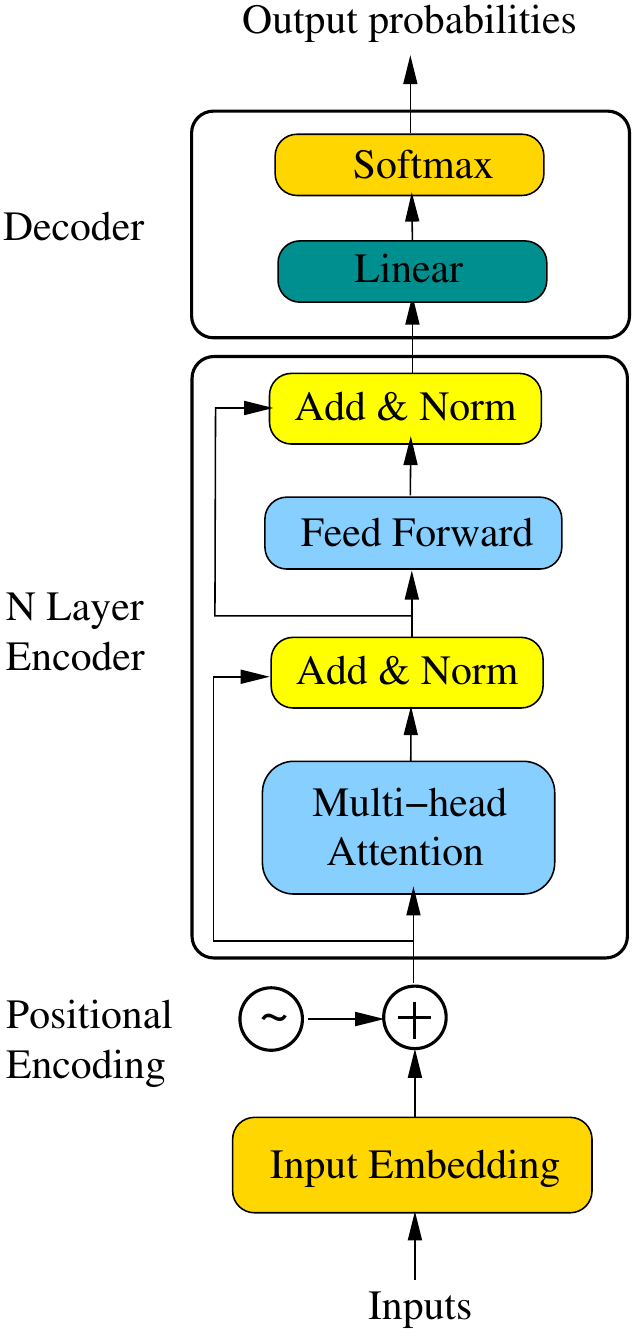}
    \caption{Transformer architecture.}
    \label{fig:trans}
\end{figure}

The BERT model consists of a transformer encoder \cite{aswani2017} as shown in Figure \ref{fig:trans}. 
The original transformer architecture uses multiple stacked self-attention layers and point-wise fully connected layers for both the encoder and decoder. However, BERT only leverages the encoder to generate hidden value representation and the original transformer decoder (for generating text in neural machine translation etc.) is replaced by a linear layer followed by a softmax layer, shown in Figure \ref{fig:trans}, both for sequential classification tasks (named entity tagging, question answering) and sentence classification tasks (sentiment classification etc.). The encoder is composed of a stack of $N=12$ or $N=24$ layers for the BERT-base and -large cases respectively. Each layer consists of two sub-layers. The first sub-layer is a multi-head self-attention mechanism, and the second sub-layer is a simple, position-wise fully connected feed-forward network. \cite{aswani2017} employs a residual connection \cite{kaiming2016} around each of the two sub-layers, followed by layer normalization \cite{jimmy2016}. That is, the output of each sub-layer is $LayerNorm(x+Sublayer(x))$, where $Sublayer(x)$ is the function implemented by the sub-layer itself. To facilitate these residual connections, all sub-layers in the model, as well as the embedding layers, produce outputs of dimension of $768$ and $1024$ for BERT-base and BERT-large, respectively. We used the same multi-head self-attention from the original paper \cite{aswani2017}. We used the same input and output representations, i.e., the embedding and positional encoding, and the same loss objective, i.e., masked LM prediction and next sentence prediction, from the BERT paper \cite{devlin2018}.

\subsection{Proposed transformer bidirectional LSTM (TRANS-BLSTM) architectures} \label{sec:trans_blstm}
Previous experiments indicated that a bidirectional LSTM model alone may not perform on par with a transformer. For example, the distillation from a transformer model to a single-layer bidirectional LSTM model \cite{tang2019} resulted in significantly lower accuracy. We also confirmed this on our experiments in Section \ref{sec:blstm}. In this paper, we hypothesize that the transformer and bidirectional LSTM may be complementary in sequence modeling. We are motivated to investigate how a bidirectional LSTM can further improve accuracy in downstream tasks relative to a classic transformer model. Figure \ref{fig:trans_blstm} shows the two proposed Transformer with Bidirectional LSTM architectures (denoted as the TRANS-BLSTM-1 and TRANS-BLSTM-2) models respectively: 
\begin{figure*}[!hbt]
    \centering
    \includegraphics[scale=0.6]{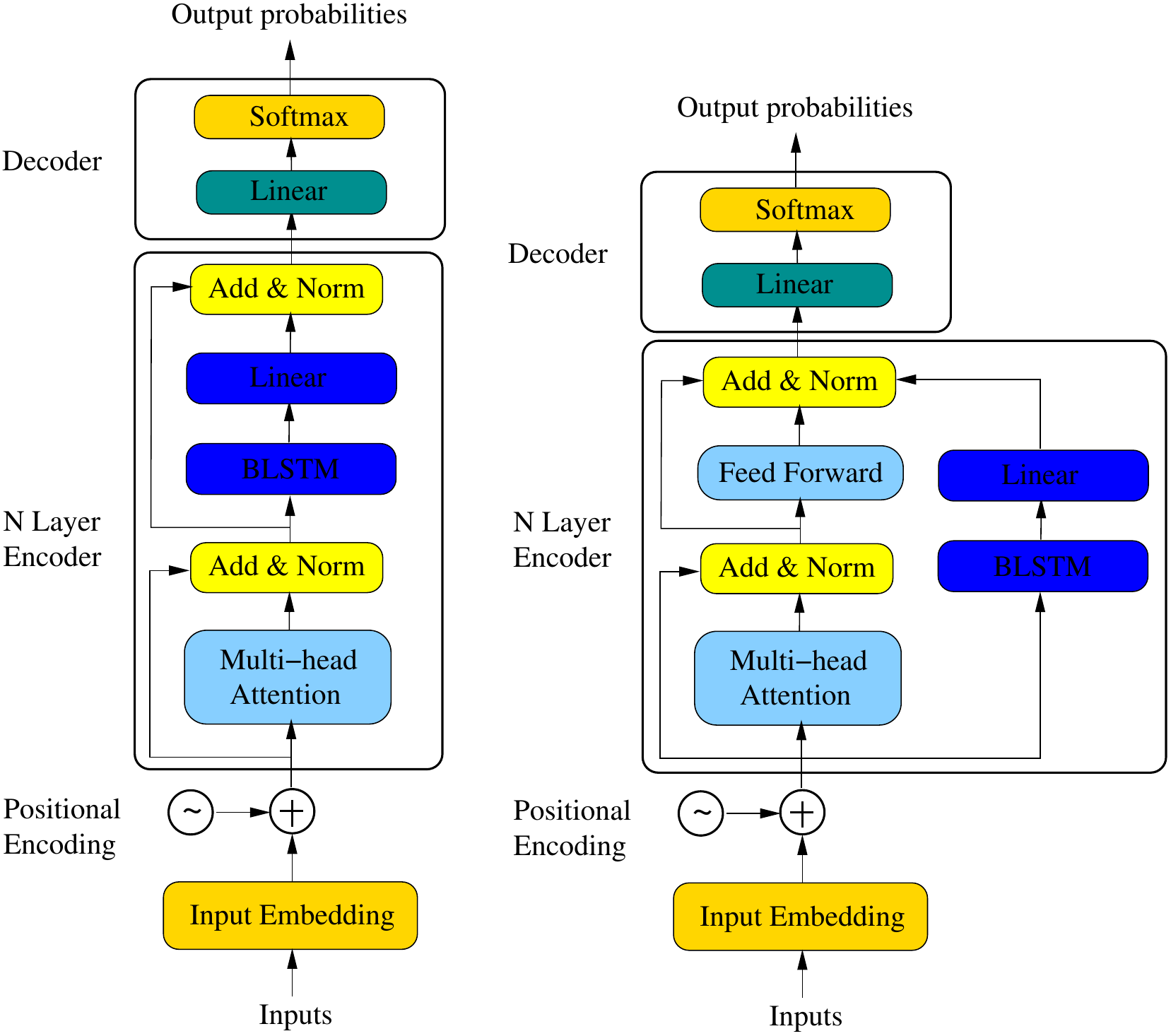}
    \caption{Two transformer with bidirectional LSTM architectures. The left one, TRANS-BLSTM-1, replaces the feedforward layer with BLSTM layer and the right , TRANS-BLSTM-2, adds a BLSTM layer in parallel.}
    \label{fig:trans_blstm}
\end{figure*}
\begin{description}
\item[TRANS-BLSTM-1] For each BERT layer, we replace the feedforward layer with a bidirectional LSTM layer.
\item[TRANS-BLSTM-2]  We add a bidirectional LSTM layer which takes the same input as the original BERT layer. The output of the bidirectional LSTM layer is summed up with the original BERT layer output (before the LayerNorm). 
\end{description}
The motivation of adding BLSTM is to integrate the self-attention and bidirectional LSTM to produce a better joint model framework (as we will see in the experiments later). We found that these two architectures lead to similar accuracy in our experiments (see Section \ref{sec:two_versions}). We thus focus on the latter (TRANS-BLSTM-2) and refer to this model as TRANS-BLSTM henceforth for simplicity. For both architectures, if we use the same number of BLSTM hidden units as in the BERT model $H$, we obtain the BLSTM output with dimension of $2H$, and we therefore need a linear layer to project the output of the BLSTM (with dimensionality $2H$) to $H$ in order to match the transformer output. Alternatively, if we set the number of BLSTM hidden units to $H/2$ (we denote this model as TRANS-BLSTM-SMALL), we need not include an additional projection layer.

\subsection{Adding bidirectional LSTM to transformer decoder}
While the above method adds bidirectional LSTM layers to a transformer encoder, we can in addition replace the linear layer with bidirectional LSTM layers in decoder.  The number of bidirectional LSTM layers is a hyper parameter to tune; we use 2 in this paper. While the bidirectional LSTM layers in encoder help the pre-training task for the masked language model and next sentence prediction task, the bidirectional LSTM in decoder may help in downstream sequential prediction tasks such as question answering. 

\subsection{Objective functions}
Following the BERT \cite{devlin2018}, we use masked language model loss and next sentence prediction (NSP) loss to train the models.

The masked LM (MLM) is often referred to as a \textit{Cloze} task in the literature \cite{taylor1953}. The encoder output, corresponding to the mask tokens, are fed into an output softmax over the vocabulary. In our experiments, we randomly mask $15\%$ of all whole word wordpiece tokens in each sequence \cite{wu2016}.

We also use the next sentence prediction loss as introduced in \cite{devlin2018} to train our models.  Specifically,
when choosing the sentences A and B for each pre-training example, $50\%$ of the time B is the actual next sentence that follows A, and $50\%$ of the time it is a random sentence from the corpus. We note that recent work \cite{yang2019,liu2019,lan2019,raffel2019} has argued that the NSP loss may not be useful in improving model accuracy. Nevertheless, we used the NSP loss in our experiments to have a fair comparison between the proposed models and the original BERT models.

\subsection{Model parameters}
Table \ref{tab:model_paras} shows the model parameter size and training speedup for TRANS/BERT (TRANS and BERT are exchangeable in this paper), TRANS-BLSTM-SMALL, and TRANS-BLSTM respectively. 
Here, the TRANS-BLSTM-SMALL and TRANS-BLSTM models are 50\% and 100\% larger than the TRANS model (base, large) respectively. Consequently, TRANS-BLSTM-SMALL and TRANS-BLSTM models require more computational resources and longer training times compared to the vanilla transformer model. The slowest-training model is the TRANS-BLSTM which is also our baseline. Models with fewer parameters can train faster. For example, the large TRANS/BERT model boasts a $2.8$ fold speedup compared to the TRANS-BLSTM large model. We note that the focus of the paper is to investigate whether a joint transformer and BLSTM architecture can further improve the performance over a transformer baseline. This is important to keep in mind because simply increasing the number of hidden units in BERT-large is not enough to positively affect accuracy \cite{lan2019} (also see Section \ref{sec:exp_squad}). 

\begin{table*}[!hbpt]
\small
\centering
    \begin{tabular}{llccccccc}
  & Model & Parameters (M) & Layers & Hidden & Embedding & Heads & Speedup\\ \hline
   & TRANS/BERT & 108M & 12 & 768 & 768 & 12 & 6.0X\\
   Base & TRANS-BLSTM-SMALL & 152M & 12 & 768 & 768 & 12 & 3.3X \\
  & TRANS-BLSTM  & 237M & 12 & 768 & 768 & 12 & 2.5X \\ 
    \hline
    Large & TRANS/BERT & 334M & 24 & 1024 & 1024 & 16 & 2.8X\\
   & TRANS-BLSTM-SMALL & 487M &  24 & 1024 & 1024 & 16  & 1.4X\\ 
   & TRANS-BLSTM & 789M &  24 & 1024 & 1024  & 16 & 1\\ \hline
\end{tabular}
    \caption{Parameter size and training speed for TRANS/BERT, TRANS-BLSTM-SMALL, and TRANS-BLSTM on base and large settings respectively.}
\label{tab:model_paras}
\end{table*} 

\section{Experiments}

\subsection{Experimental setup}

We use the same large-scale data which has been used for BERT model pre-training, the BooksCorpus (800M words) \cite{zhu2015} and English Wikipedia (2.5B words) \cite{wiki2004,devlin2018}. The two corpora consist of about 16GB of text. Following the original BERT setup \cite{devlin2018}, we format the inputs as ``[CLS] $x_1$ [SEP] $x_2$ [SEP]'', where $x_1 = x_{11}, x_{12} \ldots$ and $x_2 = x_{21}, x_{22} \ldots$ are two segments. To reduce the training memory consumption, we set the maximum input length to 256 (as opposed to 512 in the original BERT paper). We note that this setting may adversely affect the best accuracy we report in our paper \footnote{Nevertheless, our implementation of baseline BERT model obtained higher accuracy than that reported by the original BERT paper \cite{devlin2018}.}, but the relative accuracy gain by the proposed models are still valid. Similar to BERT, we use a vocabulary size of 30k with wordpiece tokenization. 

We generate the masked input from the MLM targets using unigram masking, which is denoted as \textit{whole word masking}. That is, each masking applies to a whole word at one time. We note that using n-gram masking (for example, with $n=3$) \cite{joshi2019,lan2019} with the length of each n-gram mask selected randomly can further improve the downstream task accuracy (for example, $2\%$ F1 score increase was observed on SQuAD 1.1 data set with n-gram masking and span boundary representation prediction \cite{joshi2019}). However, in the whole word masking setting, we are able to fairly compare the proposed TRANS-BLSTM models to the original BERT models. Similar to \cite{devlin2018}, the training data generator chooses $15\%$ of the token positions at random for making. If the $i$-th token is chosen, we replace the $i$-th token with (1) the [MASK] token $80\%$ of the time (2) a random token $10\%$ of the time (3) the unchanged $i$-th token $10\%$ of the time. 

The model updates use a batch size of $256$ and Adam optimizer with learning rate starting from 1e-4. Training was done on a cluster of nodes, where each node consists of 8 Nvidia Tesla V100 GPUs. We vary the node size from 1 to 8 depending on the model size. Our TRANS-BLSTM is implemented on top of Pytorch transformer repository~\footnote{https://github.com/huggingface/pytorch-transformers.}. 

\subsection{Downstream evaluation datasets}

Following the previous work \cite{devlin2018,yang2019,liu2019,lan2019}, we evaluate our models on the General Language Understanding Evaluation (GLUE) benchmark \cite{wang2018} and the Stanford Question Answering Dataset (SQuAD 1.1) \cite{rajpurkar2016}. GLUE is the General Language Understanding Evaluation benchmark consisting of a diverse collection of natural language understanding tasks. GLUE is model-agnostic and the tasks are selected to incentivize the development of general and robust NLU systems. The tasks included in GLUE are (1) Multi-Genre Natural Language Inference (MNLI) for sentence entailment classification, (2) Quora Question Pairs (QQP) for semantic equivalence classification, (3) Question Natural Language Inference (QNLI) for predicting whether the sentence in a query-sentence pair contains a correct answer, (4) Stanford Sentiment Treebank (SST-2) for sentiment analysis of movie reviews, (5) Corpus of Linguistic Acceptability (CoLA) for determining whether an English sentence is linguistically acceptable, (6) Semantic Textual Similarity (STS-B). The Stanford Question Answering Dataset (SQuAD) is a corpus consisting of 100k question/answer pairs sourced from Wikipedia.

\subsection{Bidirectional LSTM model on SQuAD dataset} \label{sec:blstm}
For the down-stream fine-tuning experiments on SQuAD 1.1 dataset, we have the following hyper-parameters for training. We set the learning rate to be 3e-5, training batch size to be 12, and the number of training epochs to be 2. 

We first run the experiment by replacing the transformer in BERT base with a bidirectional LSTM model with the same number of layers. That is, we replace the 12 transformer layers with 12 BLSTM layers. Table \ref{tab:squadBLSTM} shows the BERT base models, including the original BERT-base model in \cite{devlin2018} and our implementation, and the bidirectional LSTM model accuracy over SQuAD 1.1 development dataset. Our implementation results in a higher F1 score ($90.05\%$) compared to the original BERT-base one ($88.50\%$). This may be due to the fact that we use the whole word masking while BERT-base used partial word masking (an easier task, which may prevent from learning a better model). We found that the BLSTM model has F1 score of $83.43\%$, which is significantly worse than our TRANS/BERT baseline ($90.05\%$).

\begin{table}[!hbpt]
\centering
    \begin{tabular}{lccc}
  Model & EM & F1 \\ \hline
  BERT-base \cite{devlin2018} & 80.80 & 88.50 \\
   TRANS/BERT (ours) & 83.04 & 90.05\\
   BLSTM (ours) & 75.99 & 83.43 \\ \hline
\end{tabular}
    \caption{SQuAD development results for BERT base and the bidirectional LSTM model.}
\label{tab:squadBLSTM}
\end{table} 

\subsection{Models pre-training}
We run three pre-training experiments for base and large settings respectively. 1) BERT model training baseline (denoted as TRANS/BERT representing a transformer model or BERT), 2) TRANS-BLSTM-SMALL, with BLSTM having half of the hidden units of the transformer ($768/2=384$ on BERT base and $1024/2=512$ on BERT large) for BLSTM, and 3) TRANS-BLSTM, with BLSTM having the same hidden units as the transformer ($768$ on BERT base and $1024$ on BERT large).

\begin{figure}[!hbt]
    \centering
    \includegraphics[scale=0.6]{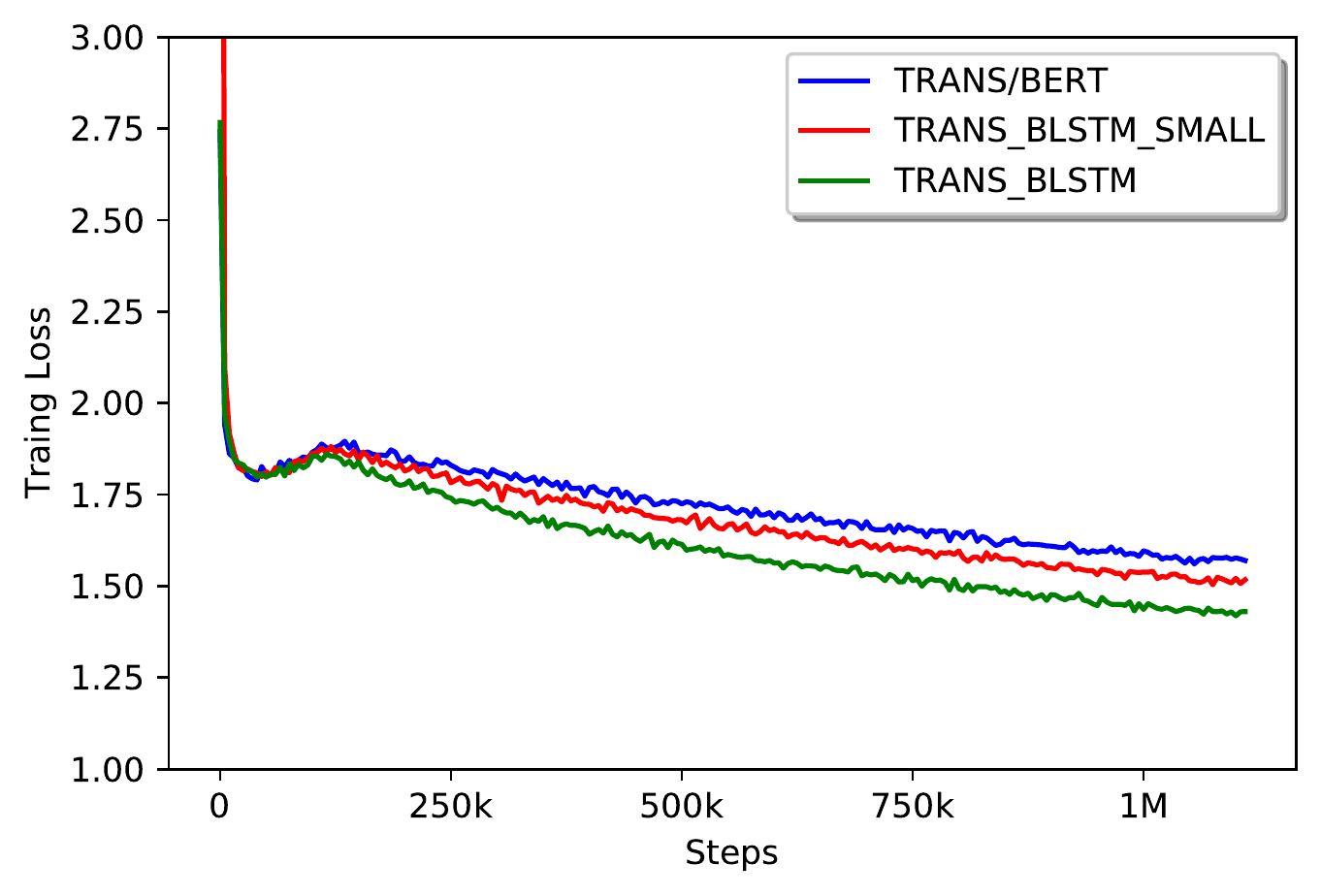}
    \caption{Training loss as a function of training steps for base TRANS/BERT, TRANS-BLSTM-SMALL, and TRANS-BLSTM models respectively.}
    \label{fig:train_loss}
\end{figure}

Figure \ref{fig:train_loss} shows the training loss for base TRANS/BERT, TRANS-BLSTM-SMALL, and TRANS-BLSTM models. As can be seen, TRANS-BLSTM-SMALL model has lower training loss than the TRANS/BERT model. TRANS-BLSTM can further decrease the training loss compared to TRANS-BLSTM-SMALL. This suggests that the proposed TRANS-BLSTM-SMALL and TRANS-BLSTM are capable of fitting the training data better than the original BERT model.

\subsection{Compare two versions of TRANS-BLSTM models} \label{sec:two_versions}
We proposed two versions of TRANS-BLSTM models in section \ref{sec:trans_blstm} (see Fig \ref{fig:trans_blstm}), with TRANS-BLSTM-1 replacing the feedforward layer with a bidirectional LSTM layer, and TRANS-BLSTM-2 adding a parallel bidirectional LSTM layer. We trained these two models and list their performance on SQuAD 1.1 development dataset in Table \ref{tab:squad_two_version}. We note that these two models lead to similar accuracy on this dataset. We will use TRANS-BLSTM-2 to report the accuracy in the rest of the experiments (denoted as TRANS-BLSTM for notational simplicity).
\begin{table}[!hbpt]
\centering
    \begin{tabular}{llccc}
  & Model & EM & F1 \\ \hline
  & TRANS-BLSTM-1  & 84.87& 91.52 \\ 
     & TRANS-BLSTM-2  & 84.75 & 91.53 \\ \hline
\end{tabular}
    \caption{SQuAD development results for two versions of  base TRANS-BLSTM models.}
\label{tab:squad_two_version}
\end{table}

\subsection{Models evaluation on SQuAD dataset} \label{sec:exp_squad}
Table \ref{tab:squad} shows the results of SQuAD dataset for TRANS/BERT, TRANS-BLSTM-SMALL and TRANS-BLSTM models for base and large settings respectively. 
\begin{table*}[!hbpt]
\centering
    \begin{tabular}{llccc}
  & Model & EM & F1 \\ \hline
  & BERT \cite{devlin2018} &  80.8 & 88.5 & \\
   & TRANS/BERT & 83.04 & 90.05\\
   & \hspace{0.1cm} + BLSTM decoder & 83.72 & 90.67 \\ 
   Base & TRANS-BLSTM-SMALL & 84.06 & 90.76\\
   & \hspace{0.1cm} + BLSTM decoder & 83.97& 90.96\\
  & TRANS-BLSTM  & \textbf{84.75} & \textbf{91.53}\\ 
  & \hspace{0.1cm} + BLSTM decoder & 84.38 & 91.25\\
    \hline
    & BERT \cite{devlin2018} &  84.1 & 90.9 & \\
    & TRANS/BERT & 85.84 & 92.34 \\
    Large & \hspace{0.1cm} + BLSTM decoder & 86.10& 92.63\\
   & TRANS-BLSTM-SMALL & 86.26 & 92.86\\ 
   & \hspace{0.1cm} + BLSTM decoder & 86.24 & 92.88  \\
   & TRANS-BLSTM & 87.72 & 93.82 \\ 
   & \hspace{0.1cm} + BLSTM decoder & \textbf{87.96} &  \textbf{94.01} \\ 
   & TRANS/BERT-48 (ours) & 85.62& 92.32 & \\ \hline
      
   & BERT xlarge \cite{lan2019} &  77.9 &  86.3 & \\
   & ALBERT xxlarge \cite{lan2019} & \textbf{88.3} & \textbf{94.1} & \\ \hline
\end{tabular}
    \caption{SQuAD development results for TRANS/BERT, TRANS-BLSTM-SMALL, and TRANS-BLSTM on base and large settings respectively.}
\label{tab:squad}
\end{table*} 
As can been see, the TRANS-BLSTM-SMALL can boost the baseline BERT model from F1 score of $90.05\%$ to $90.76\%$, and from $92.34\%$ to $92.86\%$ on base and large cases respectively. In addition, the TRANS-BLSTM can further boost accuracy on top of TRANS-BLSTM-SMALL to $91.53\%$ and $93.82\%$ on base and large respectively. The accuracy boosts suggest that the bidirectional LSTM model can add additional accuracy gain on top of the transformer models. 

Compared to adding bidirectional LSTM layers to the encoder, the addition of bidirectional LSTMs to the decoder (see +BLSTM experiments in Table \ref{tab:squad}) offers additional improvements on five out of six cases. For example, it boosts the base TRANS/BERT model F1 score of $90.05\%$ to $90.67\%$, and boosts the large TRANS-BLSTM model F1 score of $93.82\%$ to $94.01\%$. 

Table \ref{tab:squad} also shows the accuracy of original BERT models \cite{devlin2018}, which underperform our TRANS/BERT implementations, possibly due to whole word masking is used in our model training.  We also trained a TRANS/BERT-48 model, which has 48 layers (instead of the 24 layers in BERT large config) and has 638M model parameters (comparable to the model parameter size of 789M for TRANS-BLSTM). We observe that the TRANS/BERT-48 has similar accuracy as in the TRANS/BERT large model. That is, the extra depth of 24 layers does not generate additional accuracy gain compared to BERT large model. Table   
\ref{tab:squad}  also shows the the BERT xlarge model, which simply doubles the hidden units of BERT large model (ie, with $1024*2=2048$ hidden units). It has F1 score of $86.3\%$ which is significantly worse than BERT large. This suggests that simply increasing the BERT model size makes it hard to train the model, resulting in lower accuracy in this case.

Finally, we list the current state-of-the-art ALBERT model, which has F1 score of $94.1\%$ on SQuAD 1.1 development dataset. Our TRANS-BLSTM model can obtain similar accuracy to this modeling approach. 

\subsection{Model evaluation on GLUE datasets}
Following \cite{devlin2018}, we use a batch size of 32 and 3-epoch fine-tuning over the data for all GLUE tasks. For each task, we selected the best fine-tuning learning rate (among 5e-5, 4e-5, 3e-5, and 2e-5) on the development set. Additionally similar to \cite{devlin2018}, for large BERT and TRANS-BLSTM models, we found that fine-tuning was sometimes unstable on small datasets, so we ran several random restarts and selected the best model on the development set. Table \ref{tab:glue} shows the results of GLUE datasets for original BERT \cite{devlin2018}, ours TRANS/BERT, TRANS-BLSTM-SMALL and TRANS-BLSTM on base and large settings respectively. 
\begin{table*}[!hbpt]
    \centering
    \scriptsize
    \begin{tabular}{llccccccccc}
    & Model & MNLI-(m/mm) & QQP & QNLI & SST-2 &CoLA& STS-B& MRPC &  RTE  & Average  \\ \hline	
    & BERT \cite{devlin2018} & 84.6/83.4 & 71.2 &  90.5  & 93.5 & 52.1 & 85.8 &  88.9 & 66.4 & 79.6 \\
   & TRANS/BERT & 85.25/85.01 & 88.74 & 92.05 & 93.00 & 61.06 & 89.46 &  91.69 & \textbf{75.45} & 84.63 \\
    Base & TRANS-BLSTM-SMALL & 85.46/85.65 & 88.87 & \textbf{92.44} & 92.77 & 61.62 & 90.01 &  \textbf{91.71} & \textbf{75.45} & 84.77 \\
   & TRANS-BLSTM &             \textbf{86.21}/\textbf{86.36} &  \textbf{89.23}          & 92.38         & \textbf{94.26} & \textbf{62.54} & \textbf{90.67} & 91.15 &  75.40 & \textbf{85.35}\\
    \hline
  & BERT \cite{devlin2018}      & 86.7/85.9   & 72.1  &  92.7     & \textbf{94.9}  & 60.5  & 86.5  &  89.3   & 70.1  & 82.1 \\
 & TRANS/BERT                   & 87.34/87.46 & 89.16 & 93.37     & 93.92 & 62.82 & \textbf{91.03} & 89.94   & 75.30 & 85.59\\
Large & TRANS-BLSTM-SMALL & \textbf{88.19}/87.66 & \textbf{89.21} & 93.57  & 94.61 & \textbf{65.96} & 90.74 & 90.03   & 76.17 & 86.23 \\
 & TRANS-BLSTM & 88.07/\textbf{88.28}     & 88.28 & \textbf{94.08}  & 94.38 & 64.81 & 90.43 & \textbf{90.45}   & \textbf{79.78} & \textbf{86.50} \\
\hline
    \end{tabular}
    \caption{GLUE development results for TRANS/BERT, TRANS-BLSTM-SMALL and TRANS-BLSTM  on base and large settings respectively. }
\label{tab:glue}
\end{table*}
Following the BERT setting \cite{devlin2018}, we exclude the problematic WNLI set. F1 scores are reported for QQP and MRPC, Spearman correlations are reported for STS-B, and accuracy scores are reported for the other tasks. Unlike the evaluation on SQuAD dataset, we do not apply the BLSTM layer to the decoder. This is because that the tasks on GLUE are classification tasks based on the [CLS] token, and are not sequential prediction tasks (for example the SQuAD dataset) which may benefit more from including a BLSTM layer. 

We note again the accuracy discrepancy between the original BERT and our implementation of BERT, which may be due to the fact that the former uses partial word masking while the later uses whole word masking. Similar to the SQuAD results, the TRANS-BLSTM-SMALL and TRANS-BLSTM base models can improve the TRANS/BERT base model from the average GLUE score of $84.63\%$ to $84.77\%$ and $85.35\%$ respectively. In addition, the TRANS-BLSTM-SMALL and TRANS-BLSTM large models can improve the TRANS/BERT large model from the average GLUE score of $85.59\%$ to $86.23\%$ and $86.50\%$ respectively.

\section{Conclusion}
 Previous research suggested that simply increasing the hidden layer size of BERT model cannot improve the model performance. In this paper, we proposed the TRANS-BLSTM model architectures, which combine the transformer and BLSTM into one single modeling framework, leveraging the modeling capability from both transformer and BLSTM. We showed that TRANS-BLSTM models consistently lead to accuracy boost compared to transformer baselines on GLUE and SQuAD. 




\bibliography{acl2020}
\bibliographystyle{acl_natbib}

\end{document}